\documentclass[lettersize,journal]{IEEEtran}
\usepackage{amsmath,amsfonts}
\usepackage{algorithmic}
\usepackage{algorithm}
\usepackage{array}
\usepackage[caption=false,font=normalsize,labelfont=sf,textfont=sf]{subfig}
\usepackage{textcomp}
\usepackage{stfloats}
\usepackage{url}
\usepackage{verbatim}
\usepackage{graphicx}
\usepackage{cite}
\usepackage{bm}    
\usepackage{bbm}   

\usepackage[colorlinks,linkcolor=blue]{hyperref}
\usepackage{booktabs}
\usepackage{multirow}
\usepackage{color, colortbl}
\newcommand{\op}[1]{\operatorname{#1}}
\usepackage[switch]{lineno}
\begin{document}

\title{Multi-Granularity Mutual Refinement Network for Zero-Shot Learning}

\author{Ning Wang, Long Yu, Cong Hua, Guangming Zhu, Lin Mei, Syed Afaq Ali Shah, Mohammed Bennamoun, Liang Zhang* 
\thanks{Ning Wang, Long Yu, Cong Hua, Guangming Zhu, Liang Zhang are with the School of Computer Science and Technology, Xidian University, China (e-mail: ningwang@stu.xidian.edu.cn; gmzhu@xidian.edu.cn; liangzhang@xidian.edu.cn;). 

Lin Mei is with Donghai Laboratory, China, (e-mail: 13524514531@126.com).

Syed Afaq Ali Shah is with Edith Cowan University, Australia, (e-mail: afaq.shah@ecu.edu.au).

Mohammed Bennamoun is with University of Western Australia, Australia, (e-mail: mohammed.bennamoun@uwa.edu.au).}
\thanks{*Corresponding author.}
\thanks{Manuscript received January 7, 2025.}}


\markboth{Journal of \LaTeX\ Class Files,~Vol.~14, No.~8, August~2021}%
{Shell \MakeLowercase{\textit{et al.}}: A Sample Article Using IEEEtran.cls for IEEE Journals}


\maketitle

\begin{abstract}
Zero-shot learning (ZSL) aims to recognize unseen classes with zero samples by transferring semantic knowledge from seen classes. Current approaches typically correlate global visual features with semantic information (i.e., attributes) or align local visual region features with corresponding attributes to enhance visual-semantic interactions. Although effective, these methods often overlook the intrinsic interactions between local region features, which can further improve the acquisition of transferable and explicit visual features. In this paper, we propose a network named Multi-Granularity Mutual Refinement Network (Mg-MRN), which refine discriminative and transferable visual features by learning decoupled multi-granularity features and cross-granularity feature interactions. Specifically, we design a multi-granularity feature extraction module to learn region-level discriminative features through decoupled region feature mining. 
Then, a cross-granularity feature fusion module strengthens the inherent interactions between region features of varying granularities.  
This module enhances the discriminability of representations at each granularity level by integrating region representations from adjacent hierarchies, further improving ZSL recognition performance.
Extensive experiments on three popular ZSL benchmark datasets demonstrate the superiority and competitiveness of our proposed Mg-MRN method. Our code is available at \href{https://github.com/NingWang2049/Mg-MRN}{https://github.com/NingWang2049/Mg-MRN}.
\end{abstract}


\section{Introduction}

Zero-shot learning (ZSL), a task that mimics human cognitive competence, has extensively attracted increasing attention \cite{ su2023transductive, hong2023multi, ma2024lvar, wu2024re}. 
It aims to recognize images belonging to unseen categories by relying only on seen categories data. 
This is a challenging problem but particularly practical because it enables machine to recognize new objects without any previous examples, breaking free from the reliance of traditional deep learning techniques on extensive labeled datasets.
The crux of ZSL lies in learning effective visual-semantic interactions.
By utilizing auxiliary information, e.g. word2vec \cite{xian2018zero}, attributes \cite{lampert2013attribute}, and descriptions extracted from LLMs \cite{li2024visual}, it facilitates the transfer of knowledge from seen to unseen categories. 
ZSL has two settings: Conventional ZSL (CZSL) \cite{socher2013zero, akata2013label} and Generalized ZSL (GZSL) \cite{chao2016empirical, xian2018zero}. 
The former aims to predict unseen classes, and the latter can predict both seen and unseen classes. 
To track this problem, we consider two popular families of methods , i.e., generative-based methods and embedding-based methods.


Generative-based methods \cite{xian2018feature, schonfeld2019generalized, xian2019f, yu2020episode, chen2021hsva} converted ZSL into a fully supervised task by generating samples of unseen classes via generative adversarial networks (GANs) \cite{goodfellow2014generative} or variational autoencoders (VAEs) \cite{kingma2013auto}.
Although the generation-based methods achieve the seen-unseen data balance in the training process, it is difficult to overcome the performance bottleneck due to the domain gap between synthetic and real unseen samples. 
Therefore, an increasing number of studies are employing embedding-based methods \cite{akata2013label, xian2018zero, xian2016latent, annadani2018preserving, zhu2019generalized, li2023hierarchical}. 
These methods leverage attribute labels to learn a shared latent space between the visual and semantic spaces. Subsequently, they classify unseen classes within this latent space.
Some existing works \cite{zhang2017learning, chen2018zero} use global visual features related to semantic information for classification. 
Nevertheless, the global visual features are insufficient to represent fine-grained discriminative information (e.g., black-tail of Bronzed Cowbird), which is beneficial for capturing the discrepancy between different classes \cite{peng2017object}.
To solve this problem, several attention-based models \cite{xie2019attentive, zhu2019semantic, xu2020attribute, ji2018stacked, liu2019attribute} have been developed to learn region features, guided by semantic information. 
However, adopting entangled region features ignore the discriminative attribute region localization, which is necessary to improve the transferability of visual features from seen classes to unseen classes to achieve significant visual-semantic interaction.

Recently, there has been increasing interest in exploring the localization of local regions to facilitate learning more fine-grained discriminative features \cite{huynh2020fine, xie2020region, ma2023transferable}. 
For instance, attribute-guided methods \cite{chen2022transzero, wang2023bi} use semantic attention mechanisms to localize the image regions most relevant to each attribute in a given image. 
Although efforts have been made to localize ZSL attributes to local regions in images, \textit{these studies focus more on how to exploit the discriminative local regions without considering the inherent interactions between these regions}.
An object can be divided into regions that emphasize significant local regions features, and these regions also describe the whole object.
Thus, the relationship between the various local regions is interrelated and interdependent \cite{peng2017object, liu2021part}.
That is, it is often not enough to identify discriminative local regions, but also how these local regions interact amongst each other in a complementary manner.
Therefore, exploration of the relationship between different local regions can promote a more comprehensive and meaningful representation of ZSL.

To tackle this problem, we propose a Multi-granularity Mutual Refinement Network (Mg-MRN), to facilitate the interaction of features across local regions.
Specifically, inspired by the powerful ability of pyramidal convolution (PyConv) \cite{duta2020pyramidal} to capture different levels of details, we divide different local regions into different levels.
We propose a multi-granularity module (MgM) to learn multi-level feature representations without using bounding region/box annotations.
This is done by assigning features from different stages of the downsampling network to different learnable region prototypes.
Then, we introduce a mutual refinement  strategy between different granularity levels to further enhance the discriminability of different regions of each granularity level.
Moreover, considering the redundancy of high-level features relative to low-level features, we devise the spatial-channel attention to select the most relevant  information in high-level features before fusing them into low-level features.
Finally, we deploy a visual-semantic decoder module to map the region features to the attribute space under the guidance of semantic attribute information to achieve visual-semantic interaction. 
Consequently, Mg-MRN can effectively decouple discriminative local region feature and improve the transferability of knowledge through interaction between local region feature, thereby obtaining more accurate inferences for both seen and unseen categories.
Extensive experiments show that our proposed method achieves competitive performance on three ZSL benchmarks. 
Qualitative results also demonstrate that our approach can refine visual features and provide attribute-level localization. 

Our main contributions can be summarized as follows:

\begin{itemize}
    \item We propose a Multi-granularity Mutual Refinement Network (Mg-MRN) to investigate the decoupled local region feature learning and the interaction of features across local region, which encourages the transferability and discriminative attribute localization of visual features.
    \item We design a region feature mining module, which learns decoupled region-level features without using bounding region/box annotation. Furthermore, the proposed feature selection module eliminates redundancy between cross-granularity feature, enabling the improvement of the discriminative ability for the feature of the region.
    \item On three large-scale ZSL datasets, our proposed Mg-MRN outperforms state-of-the-art methods. Qualitative experiments also show that the proposed method can learn more discriminative representations.
\end{itemize}

The rest part of this paper is organized as follows. We review related works in Section II and describe the proposed networks in detail in Section III. Experimental results on three benchmark datasets are presented and analyzed in Section IV, followed by conclusions in Section V.

\section{Related Work}

\subsection{Zero-Shot Learning} 
Zero-shot learning methods can be categorized into two types: \textit{generative-based} method and \textit{embedding-based} method.
\textit{Generative-based} method solves the zero sample problem by generating samples of unseen classes.
Xian et al. \cite{xian2018feature} proposed a specialized GAN that synthesizes CNN features under the condition of category-level semantic information to address the data imbalance between seen and unseen classes.
Subsequently, Xian et al. \cite{xian2019f} developed a conditional generative-based model that combines the advantages of VAE and GANs to improve zero-shot and few-shot learning. 
Schönfeld et al.\cite{schonfeld2019generalized} introduced a modality-specific aligned VAE to construct latent features containing multimodal information related to unseen classes. 
Yu et al.\cite{yu2020episode} proposed an episodic training framework, whose core is a generative-based model that creates visual prototypes conditioned on class semantics, aligning visual and semantic information through an adversarial framework. 
Due to the heterogeneous nature of feature representations in the semantic and visual domains, Chen et al. \cite{chen2021hsva} introduced a novel Hierarchical Semantic-Visual Adaptation (HSVA) framework, which aligns the semantic and visual domains through hierarchical structure adaptation and distribution adaptation.
Tang et al. \cite{tang2021zero} proposed a Structure-Aligned Generative Adversarial Network, which generates pseudo-visual features and ensures consistency between visual and semantic spaces.
Similarly, Liu et al. \cite{liu2022zero} proposed an Attentive Region Embedding and Enhanced Semantics to improve the generation of pseudo samples for unseen classes.
Su et al. \cite{su2023dual} proposed a dual alignment strategy to align visual and semantic features to reduce the gap between generated features and real unseen features.

\textit{Embedding-based} method focus on mapping both visual features and semantic descriptions into a shared embedding space. This method allows the model to recognize unseen classes by leveraging the relationships between seen and unseen classes within this space. 
For early embedding methods, Akata et al.\cite{akata2013label}  proposed a ZSL method that utilizes properties that allow parameter sharing between seen and unseen classes as intermediate representations.
Xian et al.\cite{xian2018zero} further proposed a novel latent variable embedding model to learn compatibility between image and class embeddings in the context of zero-shot classification. 
To fully exploit the potential of semantic information associated with categories, Annadani et al.\cite{annadani2018preserving} designed objective functions to preserve the structural space formed by attributes in embedding space. However, these methods often overlook detailed differences between seen and unseen classes.
Xie et al.\cite{xie2019attentive} introduced a Attention Region Embedding Network (AREN) to emphasize on discriminative visual regions, aiming to discover semantic regions. 
Additionally, Xie et al.\cite{xie2020region} proposed a Region Graph Embedding Network (RGEN), which focuses on relationships between different local regions within the same image.
To underscore the importance of discriminative attribute localization, Chen et al. \cite{chen2022transzero} proposed the TransZero method, leveraging Transformer architecture to optimize visual features and learn attribute localization for distinctive visual embedding representation.
Liu et al. \cite{liu2023progressive} proposed a method that iteratively adapts and aligns semantic and visual features through mutual refinement, enhancing the model's ability to generalize to unseen classes.
Chen et al. \cite{chen2024progressive} proposed a framework to progressively integrate semantic guidance into visual transformers to iteratively improve feature representations.
Recently, Zhao et al. \cite{zhao2024characterizing} proposed a Transformer model for generalized zero-shot learning, which exploits hierarchical, semantic-aware part representations to improve feature disentanglement and transferability.
Although these methods are effective, they ignore the interactions between different local regions.
In this paper ,we propose the mutual refinement strategy between different granularity levels to further improve the acquisition of transferable and explicit visual features. 

\subsection{Multi-Granularity Visual Classification}
Multi-granularity features \cite{bao2022multi,chen2022cross,wang2020multi} refer to feature representations that simultaneously consider different scales or levels in images or other modal data.
Multi-granularity features can provide richer and more comprehensive information, thereby effectively solving some challenging tasks such as object detection, image classification, semantic segmentation, etc.
In visual classification tasks, early work \cite{wang2015multiple} has trained different classifiers to extract features at different levels of objects. However, this approach can only make limited use of multi-granularity information.
Du et al. \cite{du2021progressive} proposed a novel framework for fine-grained visual classification that uses a progressive training strategy and consistent block convolution to effectively fuse multi-granularity features and learn category-consistent features.
Furthermore, the PMG \cite{du2020fine} method adopted progressive training and added a jigsaw puzzle generator, forcing the model to pay attention to information at different granularity levels.
The method improves the utilization of information at different granularity levels and effectively utilizes multi-granularity features in the image.
Chen et al. \cite{chen2022label} presented a hierarchical residual network (HRN) enhanced with label relation graphs for hierarchical multi-granularity classification, which leverages hierarchical feature interactions and combinatorial loss to improve classification accuracy across different levels of granularity.
Chang et al. \cite{chang2021your} explored the challenges and methodologies in fine-grained visual classification, particularly focusing on the ambiguity and overlap between categories.
They proposed a framework to perform hierarchical feature disentanglement, with the aim of separating the adverse effects of coarse features from fine-grained features.
This approach allowed fine-grained features to contribute to coarser-level label prediction, thereby facilitating improved separation.
In addition to exploiting information at different granularities, some methods aim to study the relationship between multiple granularities or between a part and the whole object.
Peng et al.\cite{peng2017object} proposed the Object-Part Attention Model (OPAM) for weakly supervised fine-grained image classification, which integrates object-level and part-level attentions to localize objects and select discriminative parts, enhancing feature learning without the need for object or part annotations.
Additionally, Liu et al.\cite{liu2024part} introduced a Part-Object Progressive Refinement Network (POPRNet), which progressively refines discriminative and transferable semantics through the collaboration between parts and whole objects.
Inspired by these studies, our network also attempts to consider the interaction between multiple granularities while promoting interactions between regions.

\subsection{Attention Mechanism}
Attention Mechanism  \cite{vaswani2017attention} are widely used in various tasks, such as image classification, object recognition.
It is considered a significant breakthrough in the field of natural language processing\cite{devlin2018bert,radford2018improving,raffel2020exploring,dai2019transformer}.
In visual tasks \cite{zhai2022lit}, the Attention Mechanism has also been widely adopted due to its advantages in self-supervised learning and self-attention mechanisms.
Jia et al. \cite{jia2021scaling} proposed the ALIGN model, which leverages a simple dual-encoder architecture with contrastive loss to align visual and language representations, achieving state-of-the-art performance on zero-Shot learning task despite the noise in the data.
Alamri et al. \cite{alamri2021multi} introduced an attention-based model to capture and learn discriminative attributes by dividing images into small patches. 
This method enhanced zero-shot learning by effectively recognizing unseen object classes through multi-head self-attention mechanisms.
To explore discriminative regions, RGEN \cite{xie2020region} designs attention techniques to construct region graphs for transferring knowledge between different classes. This approach enhanced the ability to recognize unseen classes by leveraging both a transfer loss and a balance loss to maintain consistency between seen and unseen class outputs.
In this paper, we utilize Transformer Decoder architecture to fuse multi-granularity features and semantic information, effectively localizing the most relevant image regions for each attribute.

\begin{figure*}[htb]
    \centering
    \includegraphics[width=0.8\linewidth]{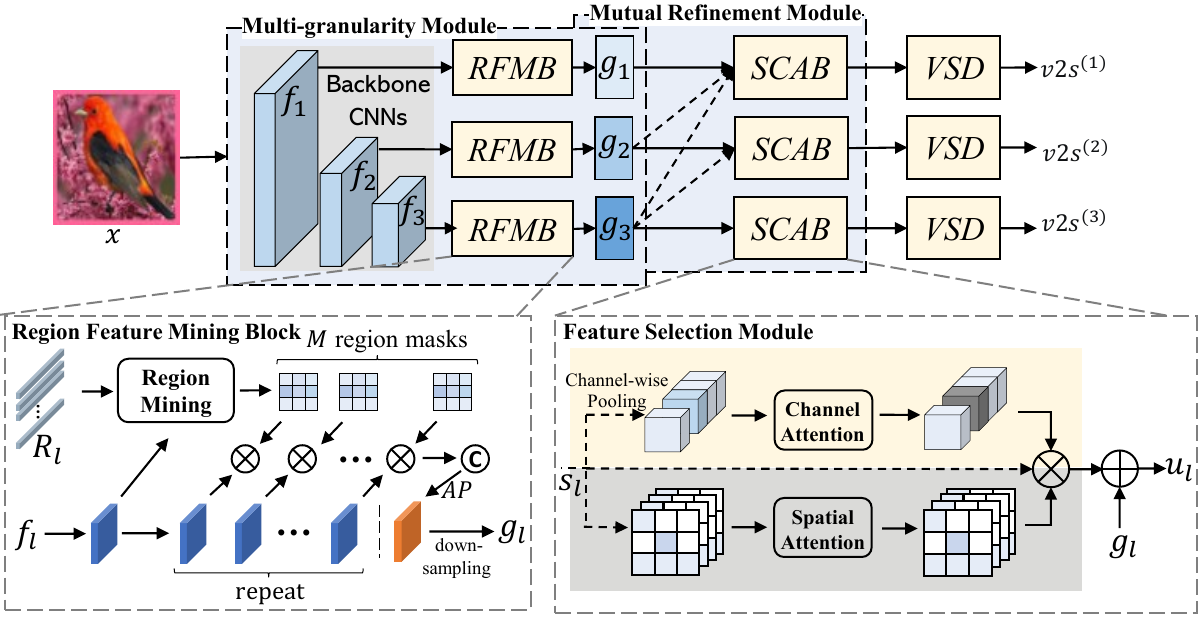}
    \caption{
    The framework of our proposed Mg-MRN. 
    It contains two innovations: (a) Multi-granularity Module, (2) Mutual Refinement Module.
    The multi-granularity module mines different grained region features from the intermediate features of the backbone network through Region Feature Mining Block (RFMB).
    The mutual refinement module exploit the  Spatial-Channel Attention Block (SCAB) to enhance the discriminability of the representation of each granularity level by integrating the region features of the adjacent hierarchies.
    The bottom left and right are the detailed RFM module and SCAB module, respectively.  Visual features are mapped to semantic space through the Visual-Semantic Decoder (VSD) module, which is detailed in Fig \ref{vsd}.
    }
    \label{overview}
\end{figure*}

\section{Proposed Method}
In this section, we introduce the Multi-granularity Mutual Refinement Network (Mg-MRN) for zero-shot learning.
An overview of the proposed Mg-MRN is illustrated in Fig \ref{overview}.
The proposed method includes three modules: the Multi-granularity Module (MgM), Mutual Refinement Module (MRM) and Visual-Semantic Decoder (VSD) Module. 
Specifically, the input image is \textbf{first} fed into MgM to extract the multi-granularity features. In this paper, we employ the ResNet-101 as backbone since it is widely used for feature extraction. After obtaining the features at each stage from the ResNet-101, region feature mining blocks are adopted to extract the specialized feature for one hierarchical level.
\textbf{Then}, the multi-granularity features are fed into MRM, which promotes mutual learning between multi-granularities to obtain a more comprehensive feature representation.
\textbf{Finally}, we use a Transformer-based VSD Module to decode visual features of each granularity guided by semantic features to enable visual-semantic interaction for ZSL classification.

\subsection{Problem Formulation}
We introduce some problem setting about zero-shot learning firstly. The main objective of Zero-Shot Learning (ZSL) is to develop a classifier capable of distinguishing the visual samples from unseen classes \( C_u \) and those from seen classes \( C_s \) in the training set, where \( C_s \cap C_u = \emptyset \). 
To achieve this goal, ZSL leverages object attribute knowledge to bridge the gap between seen and unseen classes. In the training set \( D_s \), samples \( x_i \) are drawn from seen classes \( X_s \) and are associated with corresponding semantic attributes and class labels \( y_i \) belonging to \( C_s \). 
For the CZSL task, in the test set \( D_u \), samples \( x_j \) are drawn from unseen classes \( X_u \), with the objective of predicting class labels \( c_u \) belonging to \( C_u \). In the generalized GZSL task, test samples may originate from both seen and unseen classes, with the objective of predicting class labels \( c \) belonging to \( C = C_s \cup C_u \), where \( C_s \cap C_u = \emptyset \).

\subsection{Multi-granularity Mutual Refinement Network}
\subsubsection{Multi-granularity Module}
As mentioned before, we divide local regions into different levels of granularity.
Given an input image $x$, our framework first extracts decoupled multi-granularity feature maps.
We use the CNNs backbone as the feature extractor, which has $L$ stages. 
The output feature maps from any intermediate stages are expressed as $f_l \in \mathbbm{R}^{H_l \times W_l \times C_l}$ , where $H_l$, $W_l$, $C_l$ denote the the feature map’s width, height and the number of channel at $l$-th stage, and $l = \{1, 2, …, L\}$.
Due to the limited receptive field and representational capabilities of the low stage, its output features (\textit{e.g.}, $f_1$) are forced to mine the fine-grained features (\textit{e.g.}, the local textures such as eye’s color).
On the contrary, the high-stage receptive field is larger, thus its output features (\textit{e.g.}, $f_L$) are more inclined to mine the coarse-grained features (\textit{e.g.}, the global structures such as body’s shape).
However, applying the CNN backbone network to extract multi-granularity features naively would not benefit visual-semantic interaction.
This is because the multi-granularity information extracted by different CNN stages may tend to focus on similar region.
To solve this problem, we introduce the region feature mining module (RFM) to learn region representations at different stages.
Figure \ref{overview} shows the detailed network structure of the RFM module.

For the RFM of stage $l$, we introduce a set of learnable region prototypes $R_l = \{ r_{l,m} \in \mathbbm{R}^{C_l} \}_{m=1}^{M_l}$ to discover $M_l$ different regions of $f_l$, where $r_{l,m}$ represents the $m$-th region prototype of the $l$-th stage. 
We adapt a region mining operation to produce the region masks $A_l = \{ a_{l,m} \in \mathbbm{R}^{H_l \times W_l} \}_{m=1}^{M_l}$.
Specifically, $a_{l,m}^{i,j}$ indicates the probability for the feature $f_l^{i,j}$ to be assigned to the $m$-th region $r_{l,m}$ in $R_l$, where $i,j$ index the 2D position. $a_{l,m}^{i,j}$ is computed by: 
\begin{equation}
    a_{l,m}^{i,j} = \frac{\exp(-\lVert (\bm f_l^{i,j} - \bm r_{l,m})/\sigma_{l,m} \rVert^2_2 / 2)}{\sum_{m=1}^{M_l} \exp(-\lVert (\bm f_l^{i,j} - \bm r_{l,m})/\sigma_{l,m} \rVert^2_2 / 2)},
\end{equation}
where $\sigma_{l,m}$ is a learnable smoothing factor. 
$A_l = \{ a_{l,m} \in \mathbbm{R}^{H_l \times W_l} \}_{m=1}^{M_l}$ describes the observation areas corresponding to different granularity, which helps to study the mutual promotion between multi-granularity.
Then, the features $f_l$ are repeated $M_l$ times and then weighted with the regional mask $A_l$ and further aggregated to form region representation by the average pooling:
\begin{equation}
\hat{g_l} = \frac{1}{M_l} \sum_{m=1}^{M_l} a_{l,m} \cdot f_{l,m}.
\end{equation}

Finally, we use a downsampling operation to convert the region representation $\hat{g_l}$ of each stage into features $g_l \in \mathbbm{R}^{H \times W \times C}$ with the same shape, which facilitates feature interaction between different stages in subsequent modules.

\subsubsection{Mutual Refinement Module} 
Our main motivation is to promote mutual refinement between multiple granularities, that is, fine-grained features promote the learning of coarse-grained features and vice versa.
In fact, the former already occurs in the multi-granularity feature extraction stage.
The fine-grained features of the low stage are gradually sent into the high stage through ResNet to locate the discriminative information of the global structure.
\textit{Therefore, here we aim to make coarse-grained features facilitate the learning of fine-grained features}.
Generally, coarse-grained features are redundant relative to fine-grained features because local regions are subsets of the global structure.
The features at different granularity levels correspond to different regions and different channels \cite{wang2023consistency}.
Thus, we propose a Spatial-Channel Attention Block (SCAB)  to explicitly select and fuse features from coarse-grained features that complement fine-grained features.
%
Specifically, for the SCAB of $l$-th stage, we use the features of the last $l$-1 stages (\textit{i.e.}, $l$=$L$, $L$-1, …, $L$-$l$) to enhance the features of the $l$-th stage.
We first concatenate them as:
\begin{equation}
s_l = \mathcal G_l ( \op{concat} [g_{L-l}, …, g_{L-1}, g_L]  )
\end{equation}
where $s_l \in \mathbbm{R}^{H \times W \times C}$ is the concatenated features. 
$\mathcal G_l$ refers to the fully-connected layer.
Then, we take the corresponding feature map $s_l$ as input and generate spatial attention masks $a_l^s$ and channel attention masks $a_l^c$, respectively. They can be represented as:
\begin{equation}
a_l^s = \sigma ( \boldsymbol{v}_l^s * s_l), 
\end{equation}
\begin{equation}
a_l^c = \sigma (\boldsymbol{W_2} \cdot \delta (\boldsymbol{W_1} \cdot {\rm GAP}( s_l ))),
\end{equation}
where $\sigma$ denotes the sigmoid function, $*$ refers to convolution operation and $\boldsymbol{v}_l^s$ represents convolution kernel.
$\boldsymbol{W_1}$ and $\boldsymbol{W_2}$ are the weight matrices of two FC layers.
$\rm GAP ( \cdot )$ is the global average pooling and $\delta$ refers to the ReLU function.
After that, we use the learned attention masks to weight features $s_l$, and get $u_l$:  
\begin{equation}
\hat{u_l} = s_l \cdot ( a_l^s \otimes a_l^c )
\end{equation}
where $\cdot$ represents the channel-wise product and $\otimes$ represents the Hadamard product.

Finally, we concatenate the current stage’s features $g_l$ and the selected features $\hat{u_l}$ along the channel dimension as:
\begin{equation}
u_l \leftarrow concat\left[ \delta (g_l), \delta ( \hat{u_l} ) \right] 
\end{equation}
where $\delta ( \cdot )$ is spatial flattening operation. $u_l \in \mathbbm{R}^{H \cdot W \cdot 2 \times C}$ is the augmented visual features. 

\begin{figure}[htb]
    \centering
    \includegraphics[width=0.8\linewidth]{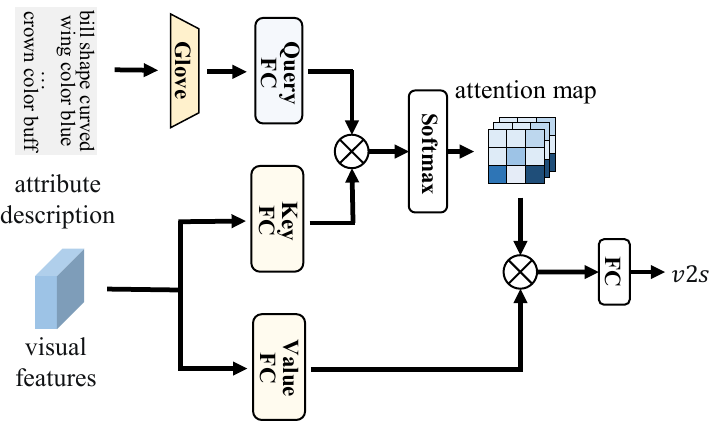}
    \caption{The architecture of Visual-Semantic Decoder Module. }
    \label{vsd}
\end{figure}

\subsubsection{Visual-Semantic Decoder} 
To achieve visual-semantic interaction, we propose a Visual-Semantic Decoder (VSD) module, which maps the enhanced visual features of each granularity into the semantic embedding space.
Specifically, our VSD first adopts the word representation model Glove \cite{pennington2014glove} to encode the attribute description into a semantic attribute vector $w_{att} \in \mathbbm{R}^{d_s \times d_{w2v}}$, where $d_s$ is the number of attribute descriptions.
Then, we use an attention network to project $w_{att}$ and $u_l$ into the latent space where visual and semantic features are aligned.
The attention network uses three fully connected (FC) layers to transform the inputs $v_A$ and $u_l$ into queries ($ q_l = w_{att} \boldsymbol{W_l^q} $ ), keys ($ k_l = u_l \boldsymbol{W_l^k} $) and values ($ v_l = u_l \boldsymbol{W_l^v}  $). $w_l^q$, $w_l^k$ and $w_l^v$ are the learnable weights.
The attention network is able to effectively localize the image regions most relevant to each attribute in a given image.
It can be defined as:
\begin{equation}
Att_{dec.}(q_l, k_l, v_l) = softmax \left( \frac{q_l k_l}{\sqrt{d_l}} \right) v_l
\end{equation}
\begin{equation}
p_l = Att_{dec.}(q_l, k_l, v_l) \boldsymbol{W_l^o} 
\end{equation}
where $d_l$ is a scaling factor, and $\boldsymbol{W_l^o}$ is a learnable weights.
Finally, following \cite{ye2023rebalanced}, we take the semantic attribute vector $w_{att}$ as support, and embed $p_l$ into the semantic attribute space to get a mapped semantic embedding:
\begin{equation}
z_l = w_{att}^T \boldsymbol{W_l} p_l
\end{equation}
where $z_l$ is the predicted semantic attribute vector, and each element represents the confidence of having a certain attribute in the input image.
$\boldsymbol{W_l}$ is an embedding matrix.
The detailed network structure of the VSD module is shown in Fig. \ref{vsd}.

\subsection{Training and Inference.} 
In order to cover both coarse-grained features and fine-grained features for zero-shot learning, we aggregate results from different stages for the overall visual-semantic interactions.
Specifically, we use semantic cross-entropy loss and attribute regression loss to optimize each stage of the model.

\textbf{Semantic Cross-Entropy Loss.}
We leverage the semantic cross-entropy loss, for a given image, to achieve the highest compatibility score with its corresponding semantic attribute vector:
\begin{equation}
\mathcal{L}_l^{SCE} = -log \frac{exp(cos \theta(z_l, \hat{z_l}))}{\sum_{\hat{c} \in C_s} exp( cos \theta(z_l, \hat{z^c}) ) }
\end{equation}
where $\hat{z_l}$ is the ground-truth semantic attribute vector. 
$cos \theta(z_l, \hat{z_l})$ define the similarity score between the L2-normalized the predicted and the ground-truth semantic attribute vector, and it can be defined as $cos \theta(z_l, \hat{z_l}) = ( {z_l}/{||z_l ||^2} )^T ( {\hat{z_l}}/{|| \hat{z_l} ||^2} ) $.

\textbf{Attribute Regression Loss.}
Following \cite{chen2022transzero}, we utilize attribute regression loss $\mathcal{L}_l^{AR}$ to encourage each stage to accurately map visual features to corresponding corresponding attribute embeddings, i.e., minimize the mean square error between the predicted and the ground-truth semantic attribute vector:
\begin{equation}
\mathcal{L}_l^{AR} = \frac{1}{N} \sum_{i=1}^{N} ||z_l - \hat{z_l} ||_2^2
\end{equation}

The total loss on all stages is:
\begin{equation}
\mathcal{L}^{total} = \sum_{l=1}^L \left( \mathcal{L}_l^{SCE} + \mathcal{L}_l^{AR} \right)
\end{equation}

In the inference stage, we first get the semantic attribute vector $z_l'$ corresponding to the test sample. Then, we  assign the best matching class $c'$ from the unseen class $c ^u$, which is formulated as:
\begin{equation}
c' = \mathop{argmax}\limits_{c \in C^u} \sum_{l=1}^L cos \theta(z_l', \hat{z_l})
\end{equation}
For the GZSL setting, the test sample may be taken from seen classes, in which $C^u$ can be replaced by $C^s \cup C^u$.

\begin {table*}[t]
    \caption {EXPERIMENTAL RESULTS (\%) UNDER THE CZSL AND GZSL SETTINGS ON PUBLIC BENCHMARKS. In ZSL, T1 represents the top-1 accuracy (\%)  for unseen classes. In  GZSL, $U$, $S$ and $H$ represent the top-1 accuracy (\%) of unseen classes, seen classes, and their harmonic mean, respectively. The best and second best results are marked in \textcolor{red}{\textbf{red}} and \textcolor{blue}{\textbf{blue}}, respectively.}
    \vspace{-10pt}
    \centering
    \scalebox{1.19}{
    \begin{tabular}[t]{l|cccc|cccc|cccc}
    \specialrule{.1em}{.00em}{.00em}
    \multirow{2}{*}{}      & \multicolumn{4}{c|}{CUB}  &  \multicolumn{4}{c|}{AwA2}   & \multicolumn{4}{c}{SUN}\\ 
    \cmidrule{2-13}

    \multirow{-2}{*}{Methods}  & \multicolumn{1}{c|}{\textit{T1}} &\textit{U} & \textit{S} & \textit{H}  & \multicolumn{1}{c|}{\textit{T1}} & \textit{U} & \textit{S} & \textit{H}  & \multicolumn{1}{c|}{\textit{T1}} & \textit{U} & \textit{S} & \textit{H} \\
    \specialrule{.1em}{.00em}{.00em}
    \textbf{Generative-based Methods}
    \\
                        f-CLSWGAN \cite{xian2018feature}       
    &\multicolumn{1}{c|}{57.3}   & 43.7           & 57.7              & 49.7
    &\multicolumn{1}{c|}{65.3}   & 56.1           & 65.5              & 60.4     
    &\multicolumn{1}{c|}{60.8}   & 42.6           & 36.6              & 39.4     
    \\
                         CADA-VAE \cite{schonfeld2019generalized}  
    &\multicolumn{1}{c|}{60.4}    & 51.6              & 53.5              & 52.4
    &\multicolumn{1}{c|}{64.0}    & 55.8              & 75.0              & 63.9
    &\multicolumn{1}{c|}{61.7}    & 47.2              & 35.7              & 40.6
    \\        
                         f-VAEGAN-D2 \cite{xian2019f}       
    &\multicolumn{1}{c|}{61.0}    & 48.4              & 60.1              & 53.6
    &\multicolumn{1}{c|}{71.1}    & 57.6              & 70.6              & 63.5     
    &\multicolumn{1}{c|}{64.7}    & 45.1              & 38.0              & 41.3   
    \\  
                        E-PGN \cite{yu2020episode}
    &\multicolumn{1}{c|}{72.4}    & 52.0              & 61.1              & 56.2 
    &\multicolumn{1}{c|}{73.4}    & 52.6              & 83.5              & 64.6    
    &\multicolumn{1}{c|}{-}       & -                 & -                 & -        
    \\    
               SAGAN$^{*}$ \cite{tang2021zero}   
    &\multicolumn{1}{c|}{58.1}    & 45.3          & 59.5              & 51.4
    &\multicolumn{1}{c|}{73.4}    & 55.9          & 84.2              & 67.2
    &\multicolumn{1}{c|}{62.9}    & 29.8          & \color{blue}{44.6}              & 35.8
    \\
    HSVA  \cite{chen2021hsva}      
    &\multicolumn{1}{c|}{62.8}    & 52.7              & 58.3              & 55.3
    &\multicolumn{1}{c|}{-}       & 59.3              & 76.6              & 66.8
    &\multicolumn{1}{c|}{63.8}    & 48.6              & 39.0              & \color{blue}{43.3}  
    \\
               GMN$^{*}$ \cite{liu2022zero}   
    &\multicolumn{1}{c|}{66.3}    & 58.8          & 60.7              & 59.7
    &\multicolumn{1}{c|}{69.5}    & 60.3          & 69.7              & 64.7
    &\multicolumn{1}{c|}{67.7}    & 46.3          & 38.4              & 42.0
    \\
    AREES$^{*}$ \cite{liu2022zero}   
    &\multicolumn{1}{c|}{65.7}    & 53.6          & 56.9              & 55.2
    &\multicolumn{1}{c|}{\color{red}{73.6}}    & 57.9          & 77.0              & 66.1
    &\multicolumn{1}{c|}{64.3}    & 51.3          & 35.9              & 42.2
    \\
    
    \specialrule{.1em}{.00em}{.00em}
    \textbf{Embedding-based Methods}
    \\
     SP-AEN \cite{chen2018zero}
    &\multicolumn{1}{c|}{55.4}    & 34.7              & 70.6              & 46.6
    &\multicolumn{1}{c|}{58.5}    & 23.3              & \color{red}{90.9}              & 37.1
    &\multicolumn{1}{c|}{59.2}    & 24.9              & 38.6              & 30.3   
    \\
                         TCN \cite{jiang2019transferable}   
    &\multicolumn{1}{c|}{59.5}    & 52.6          & 52.0              & 52.3
    &\multicolumn{1}{c|}{71.2}    & 61.2          & 65.8              & 63.4
    &\multicolumn{1}{c|}{61.5}    & 31.2          & 37.3              & 34.0
    \\
                         AGEN \cite{xie2019attentive} 
    &\multicolumn{1}{c|}{72.5}    & 63.2          & 69.0              & 66.0  
    &\multicolumn{1}{c|}{66.9}    & 54.7          & 79.1              & 64.7
    &\multicolumn{1}{c|}{60.6}    & 40.3          &32.3               & 35.9
    \\
                        SGAM \cite{zhu2019semantic}
    &\multicolumn{1}{c|}{71.0}    & 36.7          & 71.3              & 48.5 
    &\multicolumn{1}{c|}{68.8}    & 37.6          & 87.1              & 52.5 
    &\multicolumn{1}{c|}{-}       &  -            &  -                &  - 
    \\
               APN \cite{xu2020attribute}   
    &\multicolumn{1}{c|}{72.0}    & 65.3          & 69.3              & 67.2
    &\multicolumn{1}{c|}{68.4}    & 56.5          & 78.0              & 65.5
    &\multicolumn{1}{c|}{61.6}    & 41.9          & 34.0              & 37.6
    \\
    APN$^{*}$ \cite{xu2020attribute}   
    &\multicolumn{1}{c|}{75.6}    & 68.6          & 71.6              & 70.1
    &\multicolumn{1}{c|}{69.8}    & 60.1          & 86.5              & 71.0
    &\multicolumn{1}{c|}{62.6}    & 42.8          & 37.7              & 40.1
    \\
                TransZero \cite{chen2022transzero}
    &\multicolumn{1}{c|}{76.8}     & 69.3          & 68.3              & 68.8
    &\multicolumn{1}{c|}{70.1}     & 61.3          & 82.3              & 70.2
    &\multicolumn{1}{c|}{\color{blue}{65.6}}     & \color{red}{52.6}          & 33.4              & 40.8 
    \\
                           MSDN \cite{chen2022msdn}       
    &\multicolumn{1}{c|}{76.1}    & 68.7          & 67.5              & 68.1
    &\multicolumn{1}{c|}{70.1}    & 62.0          & 74.5              & 67.7    
    &\multicolumn{1}{c|}{65.8}    & \color{blue}{52.2}          & 34.2              & {41.3} 
    \\
               HAS \cite{chen2023zero}       
    &\multicolumn{1}{c|}{76.5}    & 69.6          & 74.1              & 71.8
    &\multicolumn{1}{c|}{71.4}    & 63.1          & \color{blue}{87.3}              & \color{red}{73.3}    
    &\multicolumn{1}{c|}{63.2}    & 42.8          & 38.9              & {40.8} 
    \\
            DUET \cite{chen2023duet}
    &\multicolumn{1}{c|}{72.3} & 62.9 & 72.8 & 67.5 
    &\multicolumn{1}{c|}{69.9} & 63.7 & 84.7 & \color{blue}{72.7} 
    &\multicolumn{1}{c|}{64.4} & 45.7 & \color{red}{45.8} & \color{red}{45.8}
    \\
               POPRNet \cite{liu2024part}
    &\multicolumn{1}{c|}{\color{blue}{80.4}} & \color{blue}{70.9} & \color{red}{78.8} & \color{blue}{74.6} 
    &\multicolumn{1}{c|}{69.8} & 63.7 & 83.9 & 72.4 
    &\multicolumn{1}{c|}{65.1} & 48.7 & 34.5 & 40.4
    \\    
    \specialrule{.1em}{.00em}{.00em}
    TransZSL \cite{zhao2024characterizing}
    &\multicolumn{1}{c|}{75.2} & 69.4 & 69.5 & 69.5 
    &\multicolumn{1}{c|}{71.2} & \color{red}{68.0} & 61.7 & 71.8 
    &\multicolumn{1}{c|}{\color{red}{70.3}} & 47.4 & 42.4 & 44.8
    \\    
    \specialrule{.1em}{.00em}{.00em}
               Mg-MRN (Ours)
    &\multicolumn{1}{c|}{\color{red}{81.5}} & \color{red}{74.2} & \color{blue}{77.8} &  \color{red}{75.9}
    &\multicolumn{1}{c|}{\color{blue}{71.5}} & \color{blue}{64.6} & 84.7 & \color{red}{73.3}
    &\multicolumn{1}{c|}{\color{blue}{65.6}} & 52.0 & 32.6 & 40.1
    \\
    \specialrule{.1em}{.00em}{.00em}
    \end{tabular}}
    \label{gzslperoformance}
\end {table*}

\section{Experimental results} \label{exp}
To demonstrate the effectiveness and advantages of our proposed method, we evaluate it on the metrics both for CZSL and GZSL settings on three datasets. In Sec. \ref{exp-set}, we present our experimental setup. In Sec. \ref{exp-sota}, we report our quantitative results and compare the performance with the SOTA methods. In Sec. \ref{exp-ab}, we conduct ablation studies and detailed analysis of each component. In Sec. \ref{exp-vis}, we show some visual examples to demonstrate the effectiveness of our proposed method.

\begin{table}[t]
\caption {The statistics of datasets to evaluate our proposed method.}
\centering
\scalebox{0.95}{ \begin{tabular}{c|c|c|c}
\hline
Dataset & \begin{tabular}[c]{@{}c@{}}Semantic\\ dimension\end{tabular} & \begin{tabular}[c]{@{}c@{}}Classes\\ total/seen/unseen\end{tabular} & \begin{tabular}[c]{@{}c@{}}Images\\ total/train/test unseen/seen\end{tabular} \\ \hline
CUB     & 312                                                          & 200 / 150 / 50                                                       & 11,788 / 7,057 / 2,679 / 1,764                                                 \\ \hline
 SUN     & 102                                                          & 717 / 645 / 72                                                       & 14,340 / 10,320 / 1,440 / 2,580                                                \\ \hline
AWA2    & 85                                                           & 50 / 40 / 10                                                         & 30,475 / 19,832 / 4,958 / 5,685                                                \\ \hline
\end{tabular} }
\label{dataset}
\end{table}

\subsection{Datasets and Settings} \label{exp-set}
\subsubsection{Datasets} We evaluate our proposed method on three popular ZSL benchmarks, i.e., Caltech-USCD Birds-200-2011 (CUB), SUN Attribute (SUN), Animals with Attributes2 (AWA2). 
CUB has 11,788 images of 200 bird classes (150 seen and 50 unseen) with 312 attributes. SUN has 14,340 images of 717 scene classes (645 seen and 72 unseen), including 102 attributes. AWA2 includes 37,322 images form 50 animal classes (40 seen and 10 unseen), depicted with 85 attributes. 
Table \ref{dataset} summarizes the details of the three datasets.
We split training and test sets from these datasets using the protocol of \cite{xian2018zero}, which is widely used in previous methods.

\subsubsection{Implementation details} Utilizing ResNet101 as the backbone, pretrained on ImageNet-1K, we optimized Mg-MRN using a stochastic gradient descent (SGD) optimizer with a learning rate of 0.0005, momentum of 0.9, and weight decay of 0.0001. A consistent batch size of 32 was maintained across all datasets. Our experiments were conducted on an NVIDIA RTX 3090 graphics card with 24GB of memory.

\subsubsection{Metrics} For ZSL, the Top-1 accuracy for unseen classes is calculated to assess the performance of our method. For GZSL, we consider the accuracy of both unseen and seen classes simultaneously. The harmonic mean of their accuracies (defined as H = 2 × S × U / (S + U)) is also computed as an evaluation metric. S and U denote the Top-1 accuracy of seen and unseen classes, respectively.

\subsection{Comparison with State-of-the-Arts} \label{exp-sota}
To demonstrate the superiority and competitiveness of our proposed Mg-MRN method, we compare it with the state-of-the-art methods in both CZSL and GZSL settings.
In Table \ref{exp-sota}, we compare our Mg-MRN to the recent state-of-the-art approaches. We partition them into generative-based methods and embedding-based methods. Note that our proposed is an embedding-based method.

\begin{table*}[htbp]
\caption {Ablation study of each component in Mg-MRN. MG and MRM denote the Multi-granularity Module and Mutual Refinement Module, respectively. RFMB and SCAB denote the region feature mining Block and the Spatial-Channel Attention Block, respectively.
}
\centering
\renewcommand\arraystretch{1.2}
\scalebox{0.97}{
\begin{tabular}{ll|c|cccc|cccc|cccc}
\hline
\multicolumn{2}{c|}{\multirow{2}{*}{Structure}} & \multirow{2}{*}{Loss} & \multicolumn{4}{c|}{CUB} & \multicolumn{4}{c|}{AwA2} & \multicolumn{4}{c}{SUN} \\ \cline{4-15} 
\multicolumn{2}{c|}{} &  & \textit{T1} & \textit{U} & \textit{S} & \textit{H}  &\textit{T1} & \textit{U} & \textit{S} & \textit{H} & \textit{T1} & \textit{U} & \textit{S} & \textit{H} \\ \hline
\multicolumn{2}{l|}{Baseline} & \multicolumn{1}{c|}{$L_{SCE}$} & 76.6 & 68.9 & 75.1 & 71.8 & 65.8 & 59.1 & 81.4 & 68.5 & 61.3 & 45.0 & 30.5 & 36.3 \\ \hline
\multicolumn{2}{l|}{Baseline} & \multirow{5}{*}{\begin{tabular}[c]{@{}c@{}}$L_{SCE}$\\ +$L_{AR}$\end{tabular}} & \begin{tabular}[c]{@{}c@{}}77.0\\ (+0.4)\end{tabular} & \begin{tabular}[c]{@{}c@{}}69.2\\ (+0.3)\end{tabular} & \begin{tabular}[c]{@{}c@{}}77.0\\ (+1.9)\end{tabular} & \begin{tabular}[c]{@{}c@{}}72.9\\ (+1.1)\end{tabular} & \begin{tabular}[c]{@{}c@{}}66.2\\ (+0.4)\end{tabular} & \begin{tabular}[c]{@{}c@{}}59.5\\ (+0.4)\end{tabular} & \begin{tabular}[c]{@{}c@{}}82.4\\ (+1.0)\end{tabular} & \begin{tabular}[c]{@{}c@{}}69.0\\ (+0.5)\end{tabular} & \begin{tabular}[c]{@{}c@{}}61.2\\ (-0.1)\end{tabular} & \begin{tabular}[c]{@{}c@{}}45.5\\ (+0.5)\end{tabular} & \begin{tabular}[c]{@{}c@{}}30.7\\ (+0.2)\end{tabular} & \begin{tabular}[c]{@{}c@{}}36.6\\ (+0.3)\end{tabular} \\ \cline{1-2} \cline{4-15}  
\multirow{2}{*}{Baseline+MgM} & \multicolumn{1}{|c|}{w/o RFMB} &  & \begin{tabular}[c]{@{}c@{}}80.1\\ (+3.1)\end{tabular} & \begin{tabular}[c]{@{}c@{}}74.0\\ (+3.8)\end{tabular} & \begin{tabular}[c]{@{}c@{}}75.1\\ (-1.9)\end{tabular} & \begin{tabular}[c]{@{}c@{}}74.6\\ (+1.7)\end{tabular} & \begin{tabular}[c]{@{}c@{}}68.0\\ (+1.8)\end{tabular} & \begin{tabular}[c]{@{}c@{}}61.7\\ (+2.2)\end{tabular} & \begin{tabular}[c]{@{}c@{}}82.3\\ (-0.1)\end{tabular} & \begin{tabular}[c]{@{}c@{}}70.5\\ (+1.5)\end{tabular} & \begin{tabular}[c]{@{}c@{}}62.7\\ (+1.5)\end{tabular} & \begin{tabular}[c]{@{}c@{}}49.1\\ (+3.6)\end{tabular} & \begin{tabular}[c]{@{}c@{}}30.6\\ (-0.1)\end{tabular} & \begin{tabular}[c]{@{}c@{}}37.7\\ (+1.1)\end{tabular} \\ \cline{2-2} \cline{4-15}  
 &  \multicolumn{1}{|c|}{w/ \thinspace RFMB} &  & \begin{tabular}[c]{@{}c@{}}81.1\\ (+4.1)\end{tabular} & \begin{tabular}[c]{@{}c@{}}73.4\\ (+4.2)\end{tabular} & \begin{tabular}[c]{@{}c@{}}76.6\\ (-0.4)\end{tabular} & \begin{tabular}[c]{@{}c@{}}75.0\\ (+2.1)\end{tabular} & \begin{tabular}[c]{@{}c@{}}69.6\\ (+3.4)\end{tabular} & \begin{tabular}[c]{@{}c@{}}62.7\\ (+3.2)\end{tabular} & \begin{tabular}[c]{@{}c@{}}85.2\\ (+2.8)\end{tabular} & \begin{tabular}[c]{@{}c@{}}72.3\\ (+3.3)\end{tabular} & \begin{tabular}[c]{@{}c@{}}63.6\\ (+2.4)\end{tabular} & \begin{tabular}[c]{@{}c@{}}49.2\\ (+3.7)\end{tabular} & \begin{tabular}[c]{@{}c@{}}31.6\\ (+0.9)\end{tabular} & \begin{tabular}[c]{@{}c@{}}38.5\\ (+1.9)\end{tabular} \\ \cline{1-2} \cline{4-15}  
\multirow{2}{*}{\begin{tabular}[c]{@{}c@{}}Baseline+MgM\\ +MRM (Ours)\end{tabular}} &  \multicolumn{1}{|c|}{w/o SCAB} & & \begin{tabular}[c]{@{}c@{}}80.1\\ (-1.0)\end{tabular} & \begin{tabular}[c]{@{}c@{}}73.0\\ (-0.4)\end{tabular} & \begin{tabular}[c]{@{}c@{}}76.4\\ (-0.2)\end{tabular} & \begin{tabular}[c]{@{}c@{}}74.6\\ (-0.4)\end{tabular} & \begin{tabular}[c]{@{}c@{}}69.6\\ (+0.0)\end{tabular} & \begin{tabular}[c]{@{}c@{}}62.0\\ (-0.7)\end{tabular} & \begin{tabular}[c]{@{}c@{}}85.2\\ (+0.0)\end{tabular} & \begin{tabular}[c]{@{}c@{}}71.8\\ (-0.5)\end{tabular} & \begin{tabular}[c]{@{}c@{}}64.1\\ (+0.5)\end{tabular} & \begin{tabular}[c]{@{}c@{}}49.3\\ (+0.1)\end{tabular} & \begin{tabular}[c]{@{}c@{}}31.4\\ (-0.2)\end{tabular} & \begin{tabular}[c]{@{}c@{}}38.3\\ (-0.2)\end{tabular} \\ \cline{2-2} \cline{4-15} 
&  \multicolumn{1}{|c|}{w/ \thinspace SCAB} &  & \begin{tabular}[c]{@{}c@{}}81.5\\ (+0.4)\end{tabular} & \begin{tabular}[c]{@{}c@{}}74.2\\ (+0.8)\end{tabular} & \begin{tabular}[c]{@{}c@{}}77.8\\ (+1.2)\end{tabular} & \begin{tabular}[c]{@{}c@{}}75.9\\ (+0.9)\end{tabular} & \begin{tabular}[c]{@{}c@{}}71.5\\ (+1.9)\end{tabular} & \begin{tabular}[c]{@{}c@{}}64.6\\ (+1.9)\end{tabular} & \begin{tabular}[c]{@{}c@{}}84.7\\ (-0.5)\end{tabular} & \begin{tabular}[c]{@{}c@{}}73.3\\ (+1.0)\end{tabular} & \begin{tabular}[c]{@{}c@{}}65.6\\ (+2.0)\end{tabular} & \begin{tabular}[c]{@{}c@{}}52.0\\ (+2.8)\end{tabular} & \begin{tabular}[c]{@{}c@{}}32.6\\ (+1.0)\end{tabular} & \begin{tabular}[c]{@{}c@{}}40.1\\ (+1.6)\end{tabular} \\ \hline
\end{tabular}
}
\label{table-ab}
\end{table*}

\subsubsection{Conventional Zero-Shot Learning} 
For ZSL tasks, our proposed method achieves competitive performance compared to the state-of-the-art approaches on the three datasets.
As shown in Table \ref{gzslperoformance}, our method performs better than other methods on the CUB and AWA2 dataset, achieving the most advanced experimental performance. 
On the CUB dataset, our Mg-MRN achieves a new state-of-the-art performance with a top-1 accuracy of 81.5\%, which is a significant improvement of 1.1\% over the second-best.
The CUB dataset is a fine-grained dataset where the appearance of different categories is highly similar.
Classifying these categories relies on local and discriminative features among different categories, which makes recognition more challenging.
The excellent performance verifies the effective recognition capability of the discriminative part information extracted by our Mg-MRN.
Our method also achieves comparable results with top-1 accuracy of 71.5\% and 65.6\% for the AwA2 and SUN datasets, respectively.
Compared with other methods that exploit local region information (\textit{e.g.}, TransZero \cite{chen2022transzero}, POPRNet \cite{liu2024part}), our method achieves significant gains of more than 1.1\% and 1.4\% on CUB and AWA2, respectively.
\textit{This demonstrates that the visual representations learned by our Mg-MRN using the intrinsic interactions between local region features are more discriminative than the region features learned by the existing methods that exploit local region information}.

\subsubsection{Generalized Zero-Shot Learning}
For GZSL tasks, we report the performance of our method on three datasets in Table \ref{gzslperoformance}.
Our experimental results show that the accuracy of unseen classes (U) of most works is usually lower than the seen classes (S), \textit{i.e.}, $U < S$.
This can be attributed to the fact that only the seen classes are observed during training, which results in unseen samples being classified more favorably as the seen classes during testing.
In contrast, the results on the SUN dataset show a different trend ($U > S$) due to the seen classes (645) are much larger than the unseen classes (72).
Therefore, following most state-of-the-art methods, we adopt the harmonic mean $H$ as the performance evaluation metric for the GZSL task.

On the CUB dataset, our method achieves a new state-of-the-art H indicator of 75.9\%, which beats the second-best with a large margin of 1.3\%. 
We establishes a state-of-the-art performance on unseen classes with an accuracy of 74.2\%, while also achieving competitive results on seen classes.
This demonstrates that our mechanism, which promotes mutual learning across multiple granularities, can more effectively localize the image regions most relevant to each attribute in a given image. This, in turn, enhances the transfer of knowledge from seen to unseen classes.
For the AwA2 dataset, our method can achieve 64.6\% and 73.3\% on the unseen classes and H respectively, which also achieves a new state-of-the-art performance.
This advantage can be attributed to our proposed strategy that exploits the intrinsic interactions between local region features to improve the discriminability and transferability of visual features.
For the SUN dataset, our method achieves competitive performance with a top-1 accuracy of 52.0\% for unseen classes, only 0.6\% behind the best-performing TransZero method.
For the seen classes and H metric, our method performs on par with other methods, which indicates the potential of our proposed Mg-MRN for ZSL.
\textit{Overall, our method shows strong performance in both evaluation modes on the three datasets, validating the superiority and great potential of our proposed approach}.

\begin{figure*}[htbp]
    \centering
    \includegraphics[width=\linewidth]{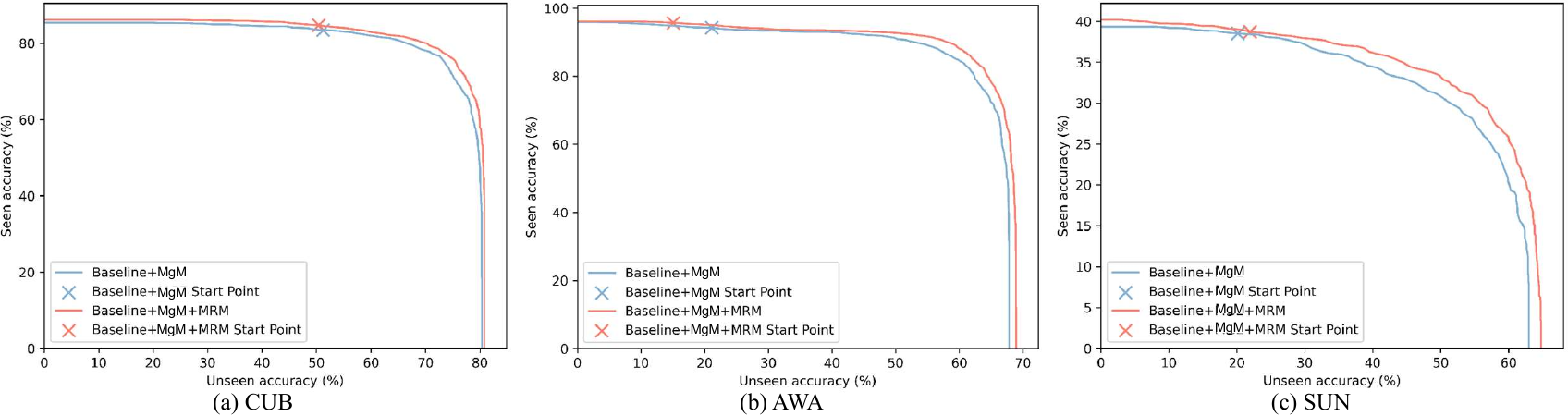}
    \caption{Visualization of Area Under Unseen-Seen Accuracy (AUSUC). After the model is equipped with the MRM, its AUSUC is significantly improved. This shows that our multi-granularity mutual refinement strategy improves the knowledge transfer from seen classes to unseen classes.}
    \label{AUSUC}
\end{figure*}

\begin{figure*}[htbp]
    \centering
    \includegraphics[width=\linewidth]{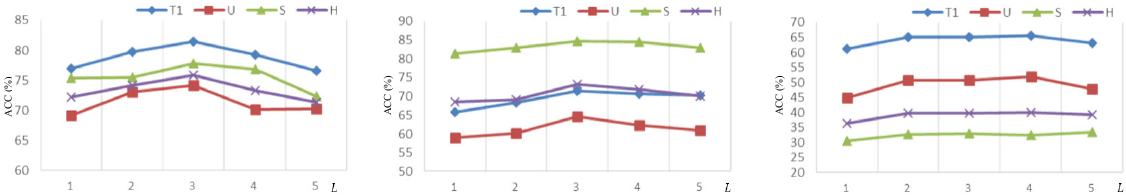}
    \includegraphics[width=\linewidth]{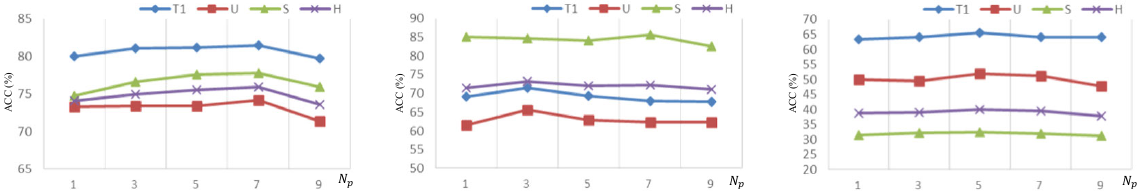}
    \caption{The effectiveness of the number of granularity level $L$ and the number of parts $N_p$ of each granularity level on (a) CUB, (b) AwA2 and (c) SUN.}
    \label{L}
\end{figure*}

\begin{figure}[htbp]
    \centering
    \includegraphics[width=0.9\linewidth]{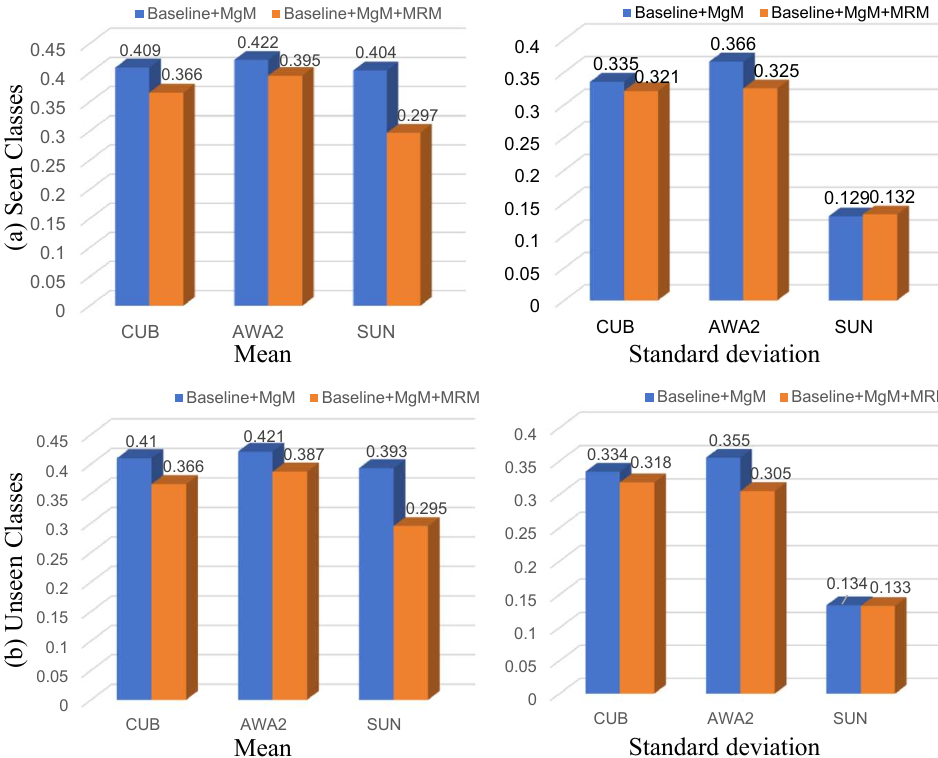}
    \caption{Mean and standard deviation of error distributions on the seen and unseen test set. This shows that our multi-granularity mutual refinement strategy produces precisely semantic predictions.}
    \label{error}
\end{figure}

\subsection{Ablation Study} \label{exp-ab}
We start with our \textit{Baseline} model which only contains a backbone and a VSD module and use the loss $L_{SCE}$ to train the baseline model.
After that, the loss $L_{AR}$ is added to the baseline model. 
From the Table \ref{table-ab},  the use of the loss $L_{AR}$ loss can contribute to the performance improvement. 
Next, we analyze the effectiveness of the proposed components, \textit{i.e.}, the multi-granularity module (MgM), the mutual refinement module (MRM) and their Hyper-Parameter.
\subsubsection{Effectiveness of MgM}
Compared to the baseline results
in row 2 of Table \ref{table-ab}, the model only using multi-granularity extraction strategy achieves a significant improvement.
This indicates that reducing the entanglement between region features can improve the transferability of visual features.
We then add the RFMB in this model, and the $T1$ and $H$ metric are further improved.
On CUB, AwA2, and SUN datasets, for ZSL, T1 improves by 4.1\%, 3.4\%, and 2.4\% compared to the baseline model , respectively.
For GZSL, H improves by 2.1\%, 3.3\%, and 1.9\% compared to the baseline model, respectively.
The results in Table \ref{table-ab} demonstrate that the RFMB can enhance the performance as it improves the discriminative ability of features at different granularity levels.

\begin{figure*}[htbp]
    \centering
    \includegraphics[width=0.95\linewidth]{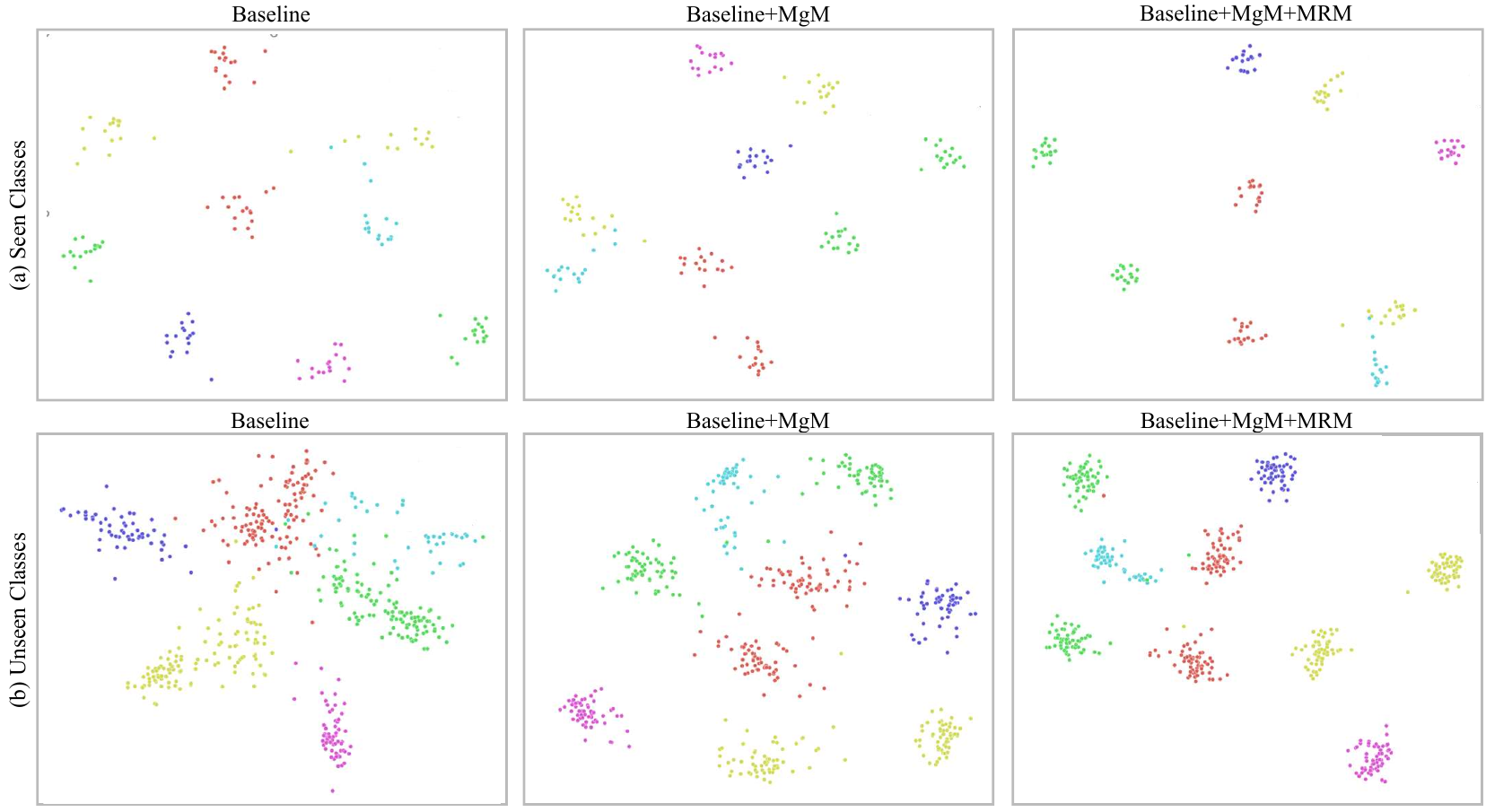}
    \caption{The t-SNE visualization for (a) seen classes and (b) unseen classes, learned by the baseline, baseline+MGR, Baseline+MgM+MRM. The 9 colors denote 9 different seen/unseen classes randomly selected from CUB dataset.}
    \label{tsne}
\end{figure*}

\subsubsection{Effectiveness of MRM}
As previously stated, the cross-granularity feature interaction can enhance the discriminability of different part regions of each granularity level.
Our proposed MRM improves the T1 of ZSL over the baseline model with MgM (\textit{i.e.}, the row 4 of Table \ref{table-ab}) by 0.4\% (CUB), 1.9\% (AWA2) and 2.0\% (SUN), respectively, and the H of GZSL by 0.9\% (CUB), 1.0\% (AWA2) and 1.6\% (SUN), respectively.
To demonstrate the effectiveness of using SCAB module, we futher evaluate the performance of our full model using different setting, i.e., with or without SCAB.
As shown in Table \ref{table-ab}, directly fusing features between different granularity levels result in no improvement or even worse performance.
This is because high-level features are redundant compared to low-level features.
The performance comparison between rows 5 and 6 in Table \ref{table-ab} can demonstrate the necessity of SCAB. 

The Area Under Unseen-Seen Accuracy (AUSUC) provides a comprehensive view of a model’s performance, which reflects the trade-off ability of ZSL between unseen accuracy and seen accuracy.
A visualization of AUSUC is shown in Fig. \ref{AUSUC}. we can see that the AUSUC of our model has been significantly improved after equipping it with MRM.
Besides, we perform a quantitative comparison of the mean and standard deviation of the error distribution on the seen and unseen classes. As shown in Fig. \ref{error}, We can observe that our proposed multi-granularity mutual refinement strategy can significantly reduce the mean and standard deviation of seen and unseen classes, which means that most semantic predictions are improved. This can be attributed to the fact that the better semantic predictions can promote learning the poor one, thereby making the overall semantic prediction better.

\subsubsection{Hyper-Parameter Analysis}
We analyzed the hyperparameters $L$ and $N_p$ on the three datasets, where $L$ is  the number of granularity level, and $N_p$ is the number of parts of each granularity level.
As shown in Fig. \ref{L}, when $L$ rises from 1 to 5, T1 and H first increase and then decrease on the three datasets. 
\textit{This phenomenon can be explained by the fact that the low-level stages of the backbone mainly focus on attribute-irrelevant features, but if forced to find attribute-relevant information from it, the overall performance will decrease}.
Thus, we set $L$ = 3 for CUB and AwA2, $L$ = 4 for SUN.
When $N_p$ rises from 1 to 9, T1 and H also increase first and then decrease on the three datasets.
\textit{The possible reason is that a small number of parts is insufficient to cover the details of attribute descriptions, while a large number of parts make the extracted region too small to keep meaningful information, leading to performance degradation}.
Thus, we set $N_p$ = 7 for CUB, $N_p$ = 3 for AWA2, and $N_p$ = 5 for SUN according to the experimental results.

\begin{figure*}[htb]
    \centering
    \includegraphics[width=0.93\linewidth]{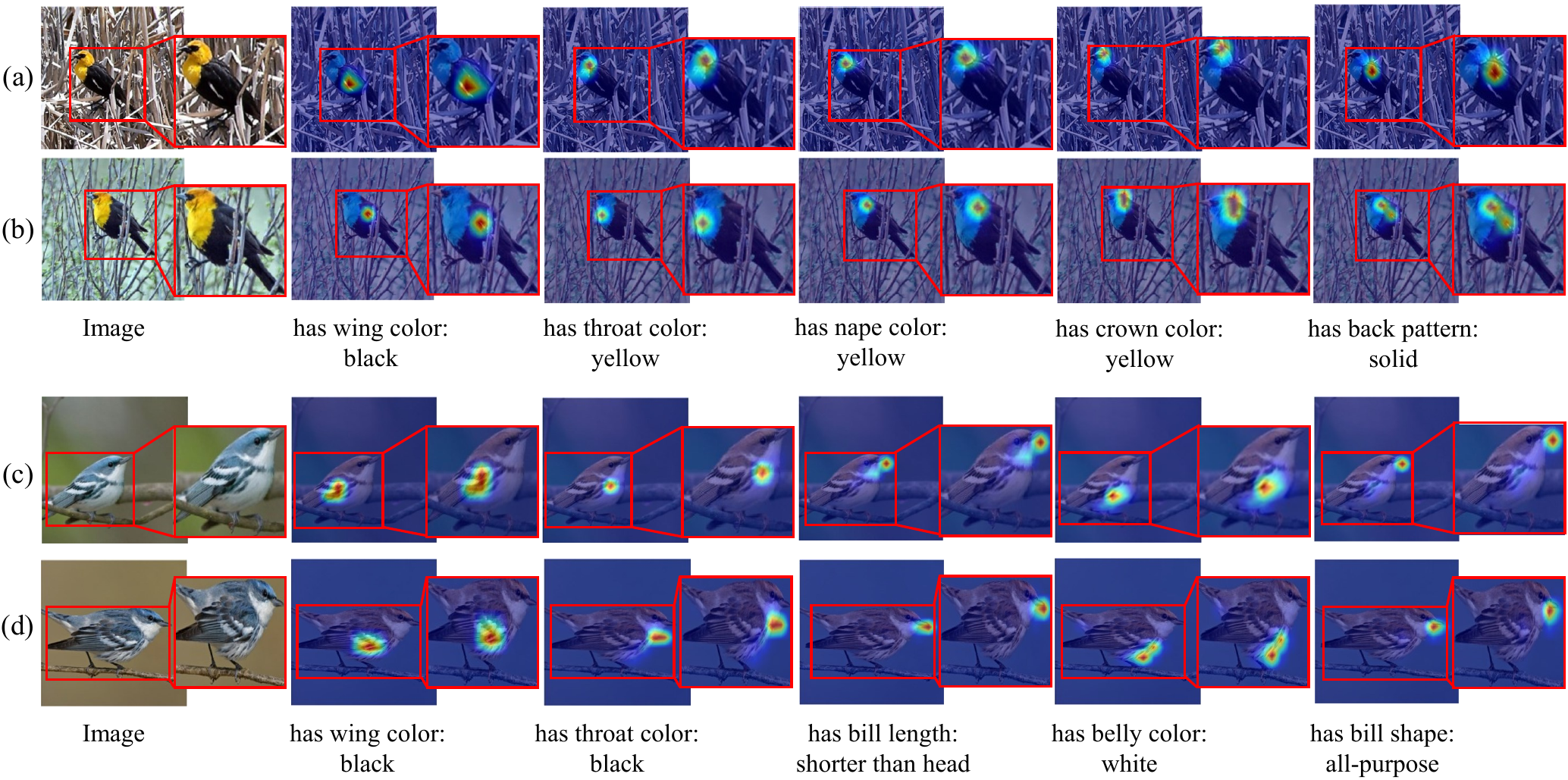}
    \caption{Visualization of attention maps produced by our method according to different semantics of unseen images on the dataset CUB. The attention map has a resolution of 14 × 14, and is reshaped into 224 × 224 to match the image size.}
    \label{vis_ours}
\end{figure*}

\begin{figure*}[htbp]
    \centering
    \includegraphics[width=0.9\linewidth]{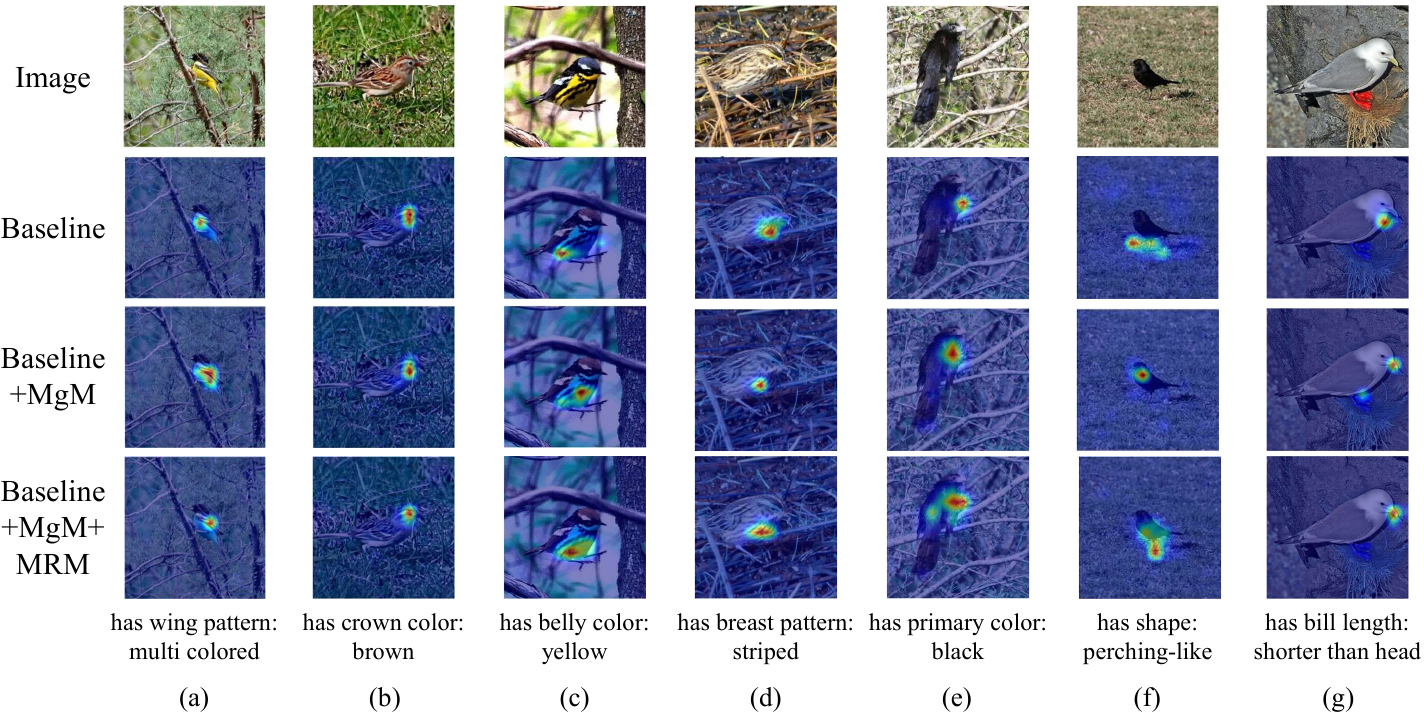}
    \caption{The effect of our proposed components  on the attention map. Our experimental settings are the same as in Fig. \ref{vis_ours}, but the model produces more accurate attention map for difficult semantics.}
    \label{vis_ab}
\end{figure*}

\subsection{Qualitative Results} \label{exp-vis}
\subsubsection{t-SNE Visualizations}
In Fig. \ref{tsne}, we show the t-SNE visualization of the visual features learned by Baseline, Baseline+MgM, and Baseline+MgM+MRM on the CUB dataset for (a) seen classes and (b) unseen classes. Compared with the visual features extracted by the Baseline model, the visual features extracted by the model with multi-granularity extraction strategy are significantly improved. When we use MRM to refine the multi-granularity visual features, the quality of the visual features of the seen and unseen classes is further enhanced. This demonstrates that our proposed method effectively combines the correlations between the different granularities features, which reduces the entanglement between different regions and improves discriminability and transferability. In addition, we also observe that MRM contributes more to the discriminability on unseen classes. This is because disentangled learning of different granularities is more conducive to cross-domain transferability, leading to better representation on unseen classes. Since the different granularities features are complementary, the comprehensive features learned by our proposed model can further improve ZSL.

\subsubsection{Attention Visualizations}
In order to qualitatively verify the effectiveness of our Mg-MRN, we visualize its attention map as shown in Fig. \ref{vis_ours}. Specifically, we fuse attention maps from all granularity levels to obtain the final attention map, which show different attention maps according to various semantics.
Obviously, our method can adaptively learn the discriminative regions corresponding to semantic information at each granularity level.
For example, when the semantics are small granularity parts (\textit{e.g.}, throat, nape, and bill), the attention of our Mg-MRN is distributed to heads of birds.
When the semantics are more larger granularity parts (\textit{e.g.}, wing, back), the attentive regions are allocated to the bird's body.
And the same semantics \textit{wing} and \textit{thrust} for distinct images are also highly specialized (columns 2-3).
The visualization result demonstrates that our model can predict the discriminative parts at each granularity level.

Furthermore, Fig. \ref{vis_ab} intuitively demonstrates the effectiveness of our proposed module.
We can see that with the addition of the proposed module, the semantics localization gets more precise.
For example, for the semantics related to bodies, the baseline model mistakenly focused on the background area.
With the help of MgM, the model corrects the out-of-focus regions.
By the region representations interaction through MRM module, fine-grained (column a, b, d, and g) semantics and coarse-grained (column c, e, and f) semantics exactly focuses on the corresponding regions, respectively.
These comparisons verify the effectiveness of our MgM module in extracting disentangled regions and our MRM module in promoting the transferability and discriminative attribute localization of visual features, respectively.

\section{Conclusion}
In this paper, we propose a Mg-MRN to explore the intrinsic interaction between region features of different granularities, which extracts enhanced discriminative and transferable visual representations for ZSL.
Mg-MRN learns disentangle multi-granularity visual features and captures part-centered region features, which alleviates visual-semantic interaction ambiguity.
To achieve the intrinsic interaction between region features of different granularities learning, a cross-granularity feature fusion module is designed to enhance the discriminability of each granularity level representation. This achieved by integrating the region representations of the adjacent hierarchies to further improve the recognition performance of ZSL
We conducted extensive experiments on CUB, AWA2 and SUN datasets to demonstrate the superior performance of our proposed method.

\bibliographystyle{IEEEtran}
\bibliography{IEEEabrv,reference}

\end{document}